\documentclass[11pt,a4paper]{article}
\usepackage[hyperref]{emnlp2020}
\usepackage{times}
\usepackage{latexsym}

\usepackage{microtype}

\aclfinalcopy 
\usepackage{sec/package}

\title{{\methodname}: Unsupervised Graph-to-Text and Text-to-Graph Generation via Cycle Training
}

\author{%
  Qipeng Guo\thanks{{} {} Equal contribution.}
  {}
  \thanks{{} {} Work done during internship at Amazon Shanghai AI Lab.}
  \\
  Fudan University\\
  \texttt{qpguo16@fudan.edu.cn} \\
   \And
   Zhijing Jin\textsuperscript{\specificthanks{1}} \\
   Amazon Shanghai AI Lab \\
   \texttt{zjin@tue.mpg.de} \\
   \And
   Xipeng Qiu \\
   Fudan University \\
   \texttt{xpqiu@fudan.edu.cn} \\
   \AND
   Weinan Zhang \\
   Shanghai Jiao Tong University \\
   \texttt{wnzhang@sjtu.edu.cn} \\
   \And
   David Wipf \\
   Amazon Shanghai AI Lab \\
   \texttt{daviwipf@amazon.com} \\
   \And
   Zheng Zhang \\
   Amazon Shanghai AI Lab \\
   \texttt{zhaz@amazon.com} \\
}

\begin{document}
\maketitle

\setcounter{footnote}{0}

\begin{abstract}
  Two important tasks at the intersection of knowledge graphs and natural language processing are graph-to-text (G2T) and text-to-graph (T2G) conversion. Due to the difficulty and high cost of data collection, the supervised data available in the two fields are usually on the magnitude of tens of thousands, for example, 18K in the WebNLG~2017 dataset after preprocessing, which is far fewer than the millions of data for other tasks such as machine translation. Consequently, deep learning models for G2T and T2G suffer largely from scarce training data.  We present \methodname{}, an unsupervised training method that can bootstrap from fully non-parallel graph and text data, and iteratively back translate between the two forms. Experiments on  WebNLG datasets show that our unsupervised model trained on the same number of data achieves performance on par with several fully supervised models. Further experiments on the non-parallel GenWiki dataset verify that our method performs the best among  unsupervised baselines. This validates our framework as an effective approach to overcome the data scarcity problem in the fields of G2T and T2G.\footnote{Our code is available at \codeurl.}
\end{abstract}

\section{Introduction}

\begin{figure}[t]
    \centering
    \includegraphics[width= \columnwidth]{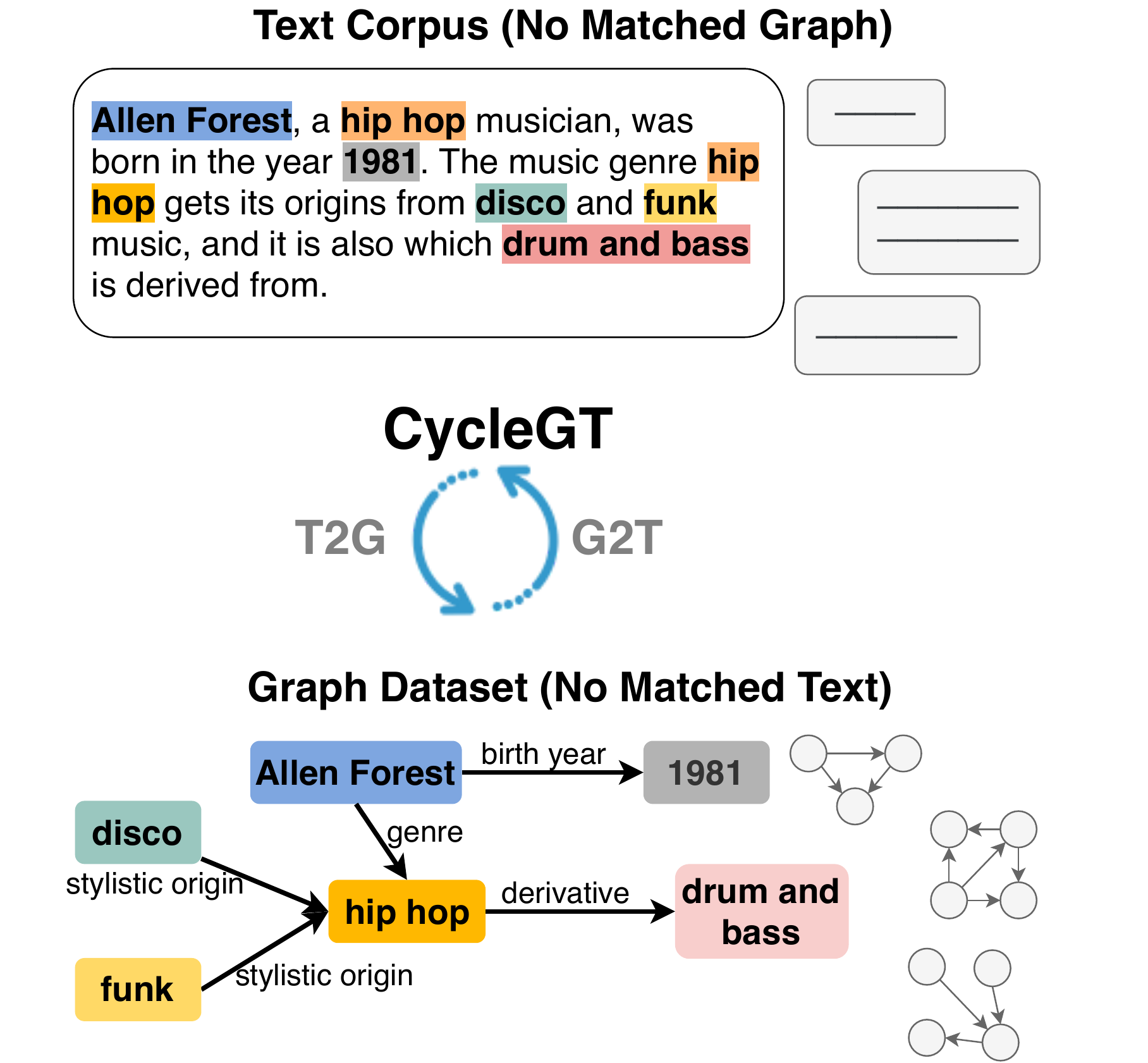}
    \caption{Given a text corpus, and a graph dataset, and no parallel (text, graph) pairs, our model \methodname{} aims to jointly learn T2G and G2T in a cycle framework.}
    \label{fig:problem_form}
\end{figure}
Knowledge graphs are a popular form of knowledge representation and central to many critical natural language processing (NLP) applications. One of the most important tasks, graph-to-text (G2T), aims to produce descriptive text that verbalizes the graphical data. For example, the knowledge graph triplet ``\textit{(Allen Forest, genre, hip hop), (Allen Forest, birth year, 1981)}'' can be verbalized as ``\textit{Allen Forest, a hip hop musician, was born in 1981}.'' This has wide real-world applications, for instance, when a digital assistant needs to translate some structured information (e.g., the properties of the restaurant) to the human user.
Another important task, text-to-graph (T2G), is to extract structures in the form of knowledge graphs from the text, so that all entities become nodes, and the relationships among entities form edges. This can serve many downstream tasks, such as information retrieval and reasoning.
The two tasks can be seen as a dual problem, as shown in Figure~\ref{fig:problem_form}.

However, most previous work has treated G2T and T2G as two separate supervised learning problems, for which the data annotation is very expensive. Therefore, both fields face the challenge of scarce parallel data. All current datasets are of a much smaller size than what is required to train the model to human-level performance. For example, the benchmark dataset WebNLG~2017 only has 18K text-graph pairs for training (after preprocessing by \citep{DBLP:conf/naacl/MoryossefGD19}), which is several magnitudes fewer than the millions of data for neural machine translation (NMT) systems, such as the 4.4M paired sentences in the WMT14 dataset for NMT \citep{luong-pham-manning:2015:EMNLP}. As a result, most previous  G2T and T2G models, which have to be trained on small datasets, display limited performance on both tasks \citep{DBLP:conf/naacl/Koncel-Kedziorski19,DBLP:conf/naacl/MoryossefGD19,DBLP:conf/semeval/LuanOH18}.

To circumvent the limitations of scarce supervised data, we  formulate both tasks in a cycle training framework, and also in the unsupervised manner with fully non-parallel datasets (as shown in Figure~\ref{fig:problem_form}). The technical difficulty lies in the different modality of text and graphs, which can be intractable in a joint learning setting.
We contribute an effective learning framework, \methodname{}, which is iteratively trained with two cycle losses.

We first validate \methodname{} on widely used WebNLG datasets, both WebNLG~2017 \citep{DBLP:conf/inlg/GardentSNP17} and WebNLG+~2020 \citep{castro-ferreira20:_2020_bilin_bidir_webnl} which is a more challenging extension of the 2017 dataset. Here \methodname{} achieves performance that is comparable to many supervised baselines.
Additionally, we also evaluate our model on the newly released GenWiki dataset \cite{genwiki2020jin} with 1.3M non-parallel text and knowledge graphs. Our model is the best out of all unsupervised baselines, improving by +11.47 BLEU over the best baseline on GenWiki\textsubscript{FINE} data and +6.26 BLEU on GenWiki\textsubscript{FULL}.

The surprisingly high performance of \methodname{} indicates that it is an effective approach to address the data scarcity problem in the fields of both G2T and T2G. 
Consequently, \methodname{}  can help pave the way for scalable, unsupervised learning, and benefit future research in both fields. 


\section{Formulation and Notations}

For unsupervised graph-to-text and text-to-graph generation,
we assume two non-parallel datasets:
\begin{itemize}[leftmargin=15px]
    \item A text corpus $D_{\mathrm{T}} = \{t_i\}^N_{i=1}$ consisting of $N$ text sequences, and
    \item A graph dataset $ D_{\mathrm{G}} = \{g_j\}^M_{j=1}$ consisting of $M$ graphs.
\end{itemize}
The constraint is that the graphs and text contain the same distribution of latent content $z$, but are different forms of surface realizations. Their
marginal log-likelihood can be formulated with the shared latent content $z$:
\begin{align}
    \log p(g) = \log \int_z p(g\mid z) p(z) dz~, \\
    \log p(t) = \log \int_z p(t\mid z) p(z) dz~.
\end{align}

Our goal is to train two models in an unsupervised manner: $\mathrm{G2T}$ that generates text based on the graph, and $\mathrm{T2G}$ that produces the graph based on text. In this vein, there is an implicit assumption that a one-to-one mapping exists, at least approximately, between text and graphs.

Denote the parameters of $\mathrm{G2T}$ as $\theta$, and parameters of $\mathrm{T2G}$ as $\varphi$. Now suppose there exists an \textit{unseen} ground truth distribution  $\mathcal{D}_{\mathrm{Pair}}$ (e.g., as in the test set), where each paired text and graph $(t, g)$ share the same content. Then the ideal objective is to maximize the log-likelihood of $\theta$ and $\varphi$ over the text and graph pairs $(t, g) \sim  \mathcal{D}_{\mathrm{Pair}}$:
\begin{align}
\begin{split}
    & \mathcal{J}(\theta, \varphi)
    \\
    & = \mathbb{E}_{(t, g)\sim  \mathcal{D}_{\mathrm{Pair}}} [\log  p ( t\mid g ;\theta) + \log  p ( g\mid t ;\varphi)]~.
\end{split}
\label{eq:overall_mll}
\end{align}

The major challenge of our task is that the ground truth distribution $\mathcal{D}_{\mathrm{Pair}}$ is not available as training data. Instead, only the text corpus $D_{\mathrm{T}}$ and graph dataset $D_{\mathrm{G}}$ are observed separately without alignment. So our solution is to approximate the learning objective in Eq.~\eqref{eq:overall_mll} through non-parallel text and graph data. The resulting method we derive, namely \methodname{}, will be introduced next. Note that we have also recently adopted the underlying \methodname{} architecture in the context of developing an alternative conditional variational autoencoder model for handling non-bijective mappings \citep{guo2020fork}.

\section{\methodname{} Development}\label{sec:method}
In this section, we will first introduce the G2T and T2G components in Section~\ref{sec:g2t} and~\ref{sec:t2g}, respectively, and then discuss the iterative back translation training strategy of \methodname{} in Section~\ref{sec:bt_n_ibt}. 

\subsection{G2T Component}\label{sec:g2t}
The model $\mathrm{G2T}: \mathcal{G} \rightarrow \mathcal{T}$ takes as input a graph $g$ and generates a text sequence ${\hat t}$ that is a sufficient description of the graph.  As pretrained models are shown to be very effective on G2T tasks \cite{ribeiro2020investigating,kale2020text}, we use T5 \cite{raffel2020exploring}, a large pretrained sequence-to-sequence model as the G2T component. Note that the model architecture of G2T is flexible in our cycle training framework, and it can also be substituted by alternatives such as the GNN-LSTM architecture as proposed in \citet{DBLP:conf/naacl/Koncel-Kedziorski19}.

Given a knowledge graph $g$, we first linearize it to a sequence $\mathrm{Seq}(g)$, using $\langle \mathrm{H}\rangle$, $\langle \mathrm{R}\rangle$, $\langle \mathrm{T}\rangle$, to denote the head, relation, and tail of a triple in the graph. As our paper mainly addresses graphs that can be verbalized in several sentences, the concerned graphs are small-scale. Hence, we linearize the graph according to the given order of triplets in the WebNLG dataset, as different linearization orders of the graphs are not a major concern, and can be handled well by the T5 model.

The finetuning of T5 aims to find the optimal parameters $\theta^*$ to correctly encode the graph sequence and decode it to the sentence. For this purpose we use maximum likelihood estimation objective
\begin{align}
\theta^* &= \argmax_{\theta } \prod_{(t,g) \sim \mathcal{D}}  p(t\mid \mathrm{Seq}(g); \theta)~.
\end{align}

\subsection{T2G Component}\label{sec:t2g}
We then introduce the other component, the T2G model. The function $\mathrm{T2G}: \mathcal{T} \rightarrow \mathcal{G}$ takes as input a text sequence $t$ and extracts its corresponding graph ${\hat g}$, whose nodes are the entities and edges are the relations between two entities. As generating both the entities and relations of the graph in a differentiable way is generally intractable, we directly use the entities if they are given, or if not given, we use an off-the-shelf entity extraction model \citep{qi2020stanza} that identifies all entities in the text with high accuracy. We denote the set of entities extracted from the text as $\mathrm{NER}(t)$. We then predict the relations between every two entities to form the edges in the graph. Note that our T2G component presented below is also flexible and can be replaced with other models.

Proceeding further, we first obtain the embeddings of every entity $v_i \in \mathrm{NER}(t) $ by average pooling the contextualized embedding of each word $w_j$ in the entity term.  This leads to
\begin{align}
    v_i &= \frac{1}{\mathrm{Len}(v_i)} \sum_{w_i \in v} \mathrm{emb}(w_j)~,
    \\
    \mathrm{emb}(w_j) &= \mathrm{enc} (w_j, w_{<j}, w_{>j})~, \nonumber
\end{align}
where $\mathrm{enc}$ encodes the embedding of $w_j$ by its preceding context $w_{<j}$ and succeeding content $w_{>j}$.

Based on the entity embeddings, we derive each edge of the graph using a multi-label classification layer $C$. $C$ takes in the two vertices of the edge and predicts the edge type, which includes the ``no-relation'' type, and the set of possible relations of entities,  leading to
\begin{align}
    e_{ij} = C(v_i, v_j).
\end{align}
T2G training aims to find the optimal parameters $\varphi^*$ that correctly encodes the text and predicts the graph via the maximum likelihood estimation objective
\begin{align}
    \varphi^* &= \argmax_{\varphi} \prod_{(t,g)\sim \mathcal{D}}  p(g\mid t; \varphi)
    \\
    &= \argmax_{\varphi} \prod_{(t,g)\sim \mathcal{D}} \prod_{i =0} ^{K} \prod_{j =0} ^{K}  p(e_{ij} \mid v_i, v_j, t; \varphi)~, \nonumber 
\end{align}
where $K=|\mathrm{NER(t)}|$ is the number of entities in text $t$.
\subsection{Cycle Training with Iterative Back Translation} \label{sec:bt_n_ibt}
In NLP, back translation \cite{sennrich-etal-2016-improving} is first proposed for machine translation where a sentence in the source language (e.g., English) should be able to first translate to the target language (e.g., French) and then again translated ``back'' to the original source language. The source sentence and the ``back-translated'' sentence should be aligned to be the same.

The essence of back translation is that a variable $\bm{x}$ and a bijective mapping function $f(\cdot)$ should satisfy $\bm{x} = f^{-1}(f(\bm{x}))$, where $f^{-1}$ is the inverse function of $f$.
In our case, G2T and T2G are inverse functions of each other, because one transforms the graph to text and the other converts text to the graph.
Specifically, we align each text with its back-translated version, and also each graph with its back-translated version via the objectives
\begin{align}
    \mathcal{L}_{\mathrm{CycT}}(\theta)
    &= \mathbb{E}_{t \in D_{\mathrm{T}}} [-\log p (t \mid \mathrm{T2G}_{\varphi}( t); \theta)]~,
    \label{eq:loss_bt1}\\
    \mathcal{L}_{\mathrm{CycG}}(\varphi)
    &= \mathbb{E}_{g \in D_{\mathrm{G}}} [-\log p (g \mid \mathrm{G2T}_{\theta}(g); \varphi)]~.
\label{eq:loss_bt2}
\end{align}
An equivalent way to interpret Eqs.~\eqref{eq:loss_bt1} and \eqref{eq:loss_bt2} is that due to the unavailability of paired text and graphs, we approximate $\mathcal{D}_{\mathrm{Pair}}$ with  $\hat D_{\mathrm{Pair}}$, a synthetic set of  text-graph pairs generated by the two models. $\hat D_{\mathrm{Pair}}$ consists of $(t, \mathrm{T2G}_{\varphi}( t))$ and $(\mathrm{G2T}_{\theta}(g), g)$ for every $t\in D_{\mathrm{T}}, g\in D_{\mathrm{G}}$, leading to


\begin{align}
\begin{split}
    \mathcal{L}_{\mathrm{CycT}}(\theta) &= \mathbb{E}_{t \in D_{\mathrm{T}}} [-\log p (t \mid \mathrm{T2G}_{\varphi}( t); \theta)]
    \\
    &= \mathbb{E}_{(t,{\hat g})  \in \hat  D_{\mathrm{Pair}}} [-\log p (t \mid {\hat g}; \theta)]~,
    \label{eq:cyc_t}
\end{split}
\\
\begin{split}
    \mathcal{L}_{\mathrm{CycG}}(\varphi) &= \mathbb{E}_{g \in D_{\mathrm{G}}} [-\log p (g \mid \mathrm{G2T}_{\theta}( t); \varphi)]
    \\
    &= \mathbb{E}_{({\hat t}, g)  \in \hat  D_{\mathrm{Pair}}} [-\log p (g \mid {\hat t}; \varphi)]~.
    \label{eq:cyc_g}
\end{split}
\end{align}
As such, to the extent that these approximations are accurate, the sum of Eqs.~\eqref{eq:loss_bt1} and \eqref{eq:loss_bt2} reasonably approximates the log likelihoods from Eq.~\eqref{eq:overall_mll}.

Note that $\mathcal{J}(\theta, \varphi) = \mathcal{L}_{\mathrm{CycT}}(\theta) + \mathcal{L}_{\mathrm{CycG}}(\varphi)$ holds when $\hat D_{\mathrm{Pair}}$ has the same distribution as $\mathcal{D}_{\mathrm{Pair}}$.
In our framework, we iteratively improve the G2T and T2G models using an iterative back translation (IBT) training scheme \cite{hoang2018iterative}, with the goal of reducing the discrepancy between the distribution of $\mathcal{D}_{\mathrm{Pair}}$ and $\hat D_{\mathrm{Pair}}$.  Specifically, we repeatedly alternate the optimization of the two cycles described by Eqs.~\eqref{eq:cyc_t} and \eqref{eq:cyc_g} over the corresponding $\theta$ or $\varphi$.

\subsection{Challenges of Cycle Training for G2T and T2G}
One of the challenges of our problem-specific use of IBT, even with the two established cycle losses above, is the \textit{non-differentiability}, which constitutes a fundamental difference between our model and the line of work represented by CycleGAN \citep{DBLP:conf/iccv/ZhuPIE17}. For our cycle losses $\mathcal{L}_{\mathrm{CycT}}$ and $\mathcal{L}_{\mathrm{CycG}}$, the intermediate model outputs are \textit{non-differentiable}. For example, in the G-Cycle (graph$\rightarrow$text$\rightarrow$graph), the intermediate text is decoded in a discrete form to natural language. Hence, the graph-to-text part $\mathrm{G2T}_{\theta}$ will not be differentiable, and the final loss can only be propagated to the latter part, text-to-graph $\mathrm{T2G}_{\varphi}$. Hence, when alternatively optimizing the two cycle losses, we first fix $\varphi$ to optimize $\theta$ for the text cycle $\mathcal{L}_{\mathrm{CycT}}$, and then fix $\theta$ to optimize $\varphi$ for the graph cycle $\mathcal{L}_{\mathrm{CycG}}$. 

Although an analogous non-differentability issue is shared by unsupervised NMT works \citep{DBLP:conf/iclr/LampleCDR18,DBLP:conf/iclr/ArtetxeLAC18}, these approaches rely on other regularization factors such as autoencoder losses, adversarial losses, and warm start strategies. In contrast, our streamlined approach \methodname{} relies solely on cycle training and the inductive biases of the T2G and G2T modules.

\section{Experiments}\label{sec:exp}
\subsection{Datasets}
\paragraph{WebNLG~2017 Dataset} 
Our first experiment uses the WebNLG~2017 dataset \citep{DBLP:conf/inlg/GardentSNP17}, which is widely used for graph-to-text generation.\footnote{WebNLG~2017 is the most appropriate dataset, because in other datasets such as relation extraction datasets \cite{walker2006ace}, the graph only contain a very small subset of the information in the text. It can be downloaded from \url{https://webnlg-challenge.loria.fr/challenge_2017/}.} Each graph consists of 2 to 7 triplets extracted from DBPedia, and the text is collected by asking crowd-source workers to describe the graphs.
The dataset includes 10 categories in the training data: Airport, Astronaut, Building, City, ComicsCharacter, Food, Monument, SportsTeam, University, and WrittenWork. There are also 5 additional unseen categories in the test data: Athlete, Artist, CelestialBody, MeansOfTransportation, and Politician.
We follow the preprocessing steps in \citet{DBLP:conf/naacl/MoryossefGD19} to obtain the text-graph pairs (with entity annotation) for 13,036 training, 1,642 validation, and 4,928 test samples.

\paragraph{WebNLG~2020 Dataset} 
We also use \methodname{} to participate in the WebNLG~2020 challenge, which is an extension of the WebNLG~2017 dataset covering broader categories. Specifically, WebNLG~2020 includes all the 15 categories of WebNLG~2017 in the training set, along with a new category, Company. Its test data also add three additional unseen categories, Film, Scientist, and MusicalWork. Overall, the dataset has 35,426 training, 4,464 validation, and 1,779 test samples.

\paragraph{GenWiki Dataset} 
Apart from the WebNLG datasets that have to be processed to fit non-parallel setting of our framework, we also evaluate with a natural non-parallel dataset, GenWiki \cite{genwiki2020jin}. GenWiki is formed from DBpedia knowledge graphs, and unpaired natural text in Wikipedia articles with the same topics as the knowledge graphs. The text has a larger vocabulary and more variety than the crowdsourced text in WebNLG. We report results on two variations of GenWiki, the full dataset GenWiki\textsubscript{FULL}, and a fine, distantly aligned version GenWiki\textsubscript{FINE}.

\subsection{Evaluation Metrics}
For G2T on WebNLG~2017 and GenWiki, we use the metrics that are mostly reported by other previous work \cite{DBLP:conf/naacl/MoryossefGD19} so that we can have head-to-head comparison with existing systems. Specifically, we adopt the metrics from \citep{DBLP:conf/naacl/MoryossefGD19}, i.e., BLEU \citep{DBLP:conf/acl/PapineniRWZ02}, Meteor \citep{DBLP:conf/acl/BanerjeeL05}, $\text{ROUGE}_{\text{L}}$ \citep{lin-2004-rouge} and CIDEr \citep{DBLP:conf/cvpr/VedantamZP15}, to measure the closeness of the
reconstructed text (model output) to the input text.\footnote{We calculate all metrics using the pycocoevalcap tool (https://github.com/salaniz/pycocoevalcap).} Briefly, they measure the n-gram precision, recall, or F-scores between the model outputs and the (ground-truth) references.  In contrast, for T2G evaluation on WebNLG 2017, we use the micro and macro F1 scores of the relation types of the edges, following the standard practice in relation extraction~\citep{miwa-bansal-2016-end,DBLP:conf/acl/ZhouSTQLHX16}.

Turning to WebNLG~2020, we report most major metrics listed on the leaderboard  \cite{moussalem20:_gener_bench_framew_text_gener}. For G2T, the metrics include BLEU, METEOR, chrF++~\cite{popovic2017chrf}, TER \cite{snover2006study},
BERT\textsubscript{F1} \cite{bertscore2020zhang}, and BLUERT \cite{sellam2020bleurt}. 

\subsection{Comparison Systems for WebNLG~2017}\label{sec:appd_baseline}
\begin{table*}[ht]
    \centering
    \small
    \begin{tabular}{lBBBBBBB}
        \toprule
        & \multicolumn{4}{c}{\textbf{G2T Performance}} & \multicolumn{2}{c}{\textbf{T2G Performance}} \\
        & BLEU & METEOR & $\text{ROUGE}_{\text{L}}$ & CIDEr & Micro F1 & Macro F1 \\
        \hline
        \multicolumn{5}{l}{\textbf{{Supervised Models (G2T Only) }}}
        \\
        \quad {Melbourne (introduced in \citep{DBLP:conf/acl/GardentSNP17})} & 45.0 & 0.376 & 63.5 & 2.81 & -- & -- \\
        \quad {StrongNeural
        \citep{DBLP:conf/naacl/MoryossefGD19}
        } & 46.5 & 0.392 & 65.4 & 2.87 & --&--\\
        \quad {BestPlan
        \citep{DBLP:conf/naacl/MoryossefGD19}
        } & 47.4 & 0.391 & 63.1 & 2.69 & -- & -- \\
        \quad {GraphWriter
        \citep{DBLP:conf/naacl/Koncel-Kedziorski19}
        } & 45.8 & 0.356 & 68.6 & 3.14 & -- & -- \\
        \quad {Seg\&Align \cite{shen2020neural}} & 46.1 & 0.398 & 65.4 & 2.64 & & \\
        \quad {Planner
        \citep{zhao-etal-2020-bridging}
        } & 52.8 & 0.450 & -- & -- & -- & -- \\
        \quad {T5-Large
        \citep{kale2020text}
        } & 57.1 & 0.440 & -- & -- & -- & -- \\
        \quad {T5-Large
        \citep{ribeiro2020investigating}
        } & 59.3 & 0.440 & -- & -- & -- & -- \\
        \quad {Supervised G2T by T5-Base (Our Implementation)} & 56.4 & 0.445 & 69.1 & 3.86 & -- & -- \\
        \multicolumn{5}{l}{\textbf{{Supervised Models (T2G Only) }}} \\
        \quad {Supervised T2G (Our Implementation)} & -- & -- & -- & -- & 60.6 & 50.7 \\
        \quad {OnePass
        \citep{DBLP:conf/acl/WangTYCWXGP19}
        } & -- & -- & -- & -- & 66.2 & 52.2 \\
        \multicolumn{5}{l}{\textbf{{Unsupervised Models}}} \\
        \quad RuleBased \cite{schmitt2019unsupervised} & 18.3 & 0.336 & -- & -- & 0.0 & 0.0 \\
        \quad GT-BT \cite{schmitt2019unsupervised} & 37.7 & 0.355 & -- & -- & 39.1 & -- \\
        \quad \textbf{\methodname{} (Unsup.)} & 55.5 & 0.437 & 68.3 & 3.81 & 58.4 & 46.4 \\
        \bottomrule
    \end{tabular}
    \caption{T2G and G2T performance of supervised and unsupervised models on WebNLG~2017 data.}
    \label{tab:res}
\end{table*}
\begin{table*}[ht]
\centering
\resizebox{\textwidth}{!}{
\begin{tabular}{@{\extracolsep{4pt}}lcccccccc@{}}
\toprule
    & \multicolumn{4}{c}{\textbf{GenWiki$_{\textsc{FINE}}$}} & \multicolumn{4}{c}{\textbf{GenWiki$_{\textsc{FULL}}$}}
    \\ 
    \cline{2-5} \cline{6-9}
    & BLEU & METEOR & ROUGE$_{\text{L}}$ & CIDEr & BLEU & METEOR & ROUGE$_{\text{L}}$ & CIDEr
    \\ 
    \hline
    \textbf{Rule-Based} \cite{schmitt2019unsupervised} & 13.45 & 30.72 & 40.93 & 1.26 & 13.45 & 30.72 & 40.93 & 1.26 \\
    \textbf{DirectTransfer} \cite{genwiki2020jin} & 13.89 & 25.76 & 39.75 & 1.26 & 13.89 & 25.76 & 39.75 & 1.26 \\
    \textbf{NoisySupervised} \cite{genwiki2020jin} & 30.12 & 28.12 & 56.96 & 2.52 & 35.03 & 33.45 & 58.14 & 2.63 \\
    \textbf{\methodname{}} & 41.59 & 35.72 & 63.31 & 3.57 & 41.29 & 35.39 & 63.73 & 3.53\\
\bottomrule
\end{tabular}
}
\caption{The performance of our \methodname{} and three unsupervised baselines on GenWiki$_{\textsc{FINE}}$ and GenWiki$_{\textsc{FULL}}$.}
\label{tab:genwiki}
\end{table*}{}
\begin{table}[ht]
    \centering
    \small
    \resizebox{\columnwidth}{!}{%
    \begin{tabular}{lDDDDD}
    \toprule
        & BLEU & METEOR & CHRF++ & TER & BLUERT \\ \hline
ID18 & 53.98 & 0.417&   0.690&  0.406 & 0.62 \\
ID30    & 53.54 & 0.414 & 0.688 & 0.416 & 0.61 \\
ID30\_1 & 52.07 & 0.413 & 0.685 & 0.444 & 0.58 \\
ID34    & 52.67 & 0.413 & 0.686 & 0.423 & 0.6 \\
ID5 & 51.74 & 0.411 & 0.679 & 0.435 & 0.6 \\
ID35    & 51.59 & 0.409 & 0.681 & 0.431 & 0.59 \\
ID23    & 51.74 & 0.403 & 0.669 & 0.417 & 0.61 \\
ID2 & 50.34 & 0.398 & 0.666 & 0.435 & 0.57 \\
ID15    & 40.73 & 0.393 & 0.646 & 0.511 & 0.45 \\
\textbf{\methodname{} (Unsup.)}  & 44.56 & 0.387 & 0.637 & 0.479 & 0.54 \\
ID12 &   40.29 &   0.386&   0.634&   0.504&   0.45 \\
ID11 &   39.84 &   0.384&   0.632&   0.517&   0.43 \\
ID4 &50.93 &   0.384&   0.636&   0.454&   0.54 \\
ID26 &   50.43 &   0.382&   0.637&   0.439&   0.57 \\
Official Baseline1 &   40.57 &   0.373&   0.621&   0.517&   0.47 \\
ID17 &   39.55 &   0.372&   0.613&   0.536&   0.37 \\
ID31\_2 & 41.03 &   0.367&   0.608&   0.522&   0.39 \\
Official Baseline2 & 37.89 &   0.364&   0.606&   0.553&   0.42 \\
ID21 &   31.98 &   0.350&   0.545&   0.629&   0.4 \\
ID13\_11 &38.37 &   0.343&   0.584&   0.587&   0.33 \\
ID14 &   39.12 &   0.337&   0.579&   0.564&   0.37 \\
ID13 &   38.2 &   0.335&   0.571&   0.577&   0.29 \\
ID13\_3 & 39.19 &   0.334&   0.569&   0.565&   0.28 \\
ID13\_4 & 38.85 &   0.332&   0.569&   0.573&   0.3 \\
ID13\_2 & 38.01 &   0.331&   0.565&   0.583&   0.27 \\
ID13\_6 & 37.96 &   0.331&   0.566&   0.580&   0.28 \\
ID10 &   22.84 &   0.326&   0.534&   0.696&   -0.03 \\
ID13\_7 & 36.07 &   0.324&   0.555&   0.599&   0.24 \\
ID13\_8 & 36.6 &   0.322&   0.554&   0.594&   0.25 \\
ID20 &   31.26 &   0.316&   0.542&   0.659&   0.31 \\
ID26\_1 & 27.5 &   0.305&   0.519&   0.846&   0.03 \\
ID13\_9 & 28.71 &   0.243&   0.448&   0.689&   -0.09 \\
ID13\_10 &30.79 &   0.239&   0.439&   0.677&   -0.09 \\
ID13\_5 & 23.08 &   0.225&   0.421&   0.743&   -0.19 \\
ID17\_1 & 24.45 &   0.223&   0.425&   0.739&   -0.22 \\
\bottomrule
    \end{tabular}
    }
    \caption{Leaderboard of the submitted systems and two official baselines (Baseline1 and Baseline2) in the WebNLG~2020 English RDF-to-text challenge as \textit{ranked by METEOR}. With all metrics, larger is better, with the exception of TER where lower is better. }
    \label{tab:webnlg20_g2t}
\end{table}

\paragraph{Unsupervised Baselines}
As cycle training models are unsupervised learning methods, we include the following unsupervised baselines:
\begin{itemize}[nolistsep]
    \item \textit{RuleBased} \citep{schmitt2019unsupervised}: This baseline involves simply iterating through the graph and concatenating the text of each triplet.
    \item \textit{Graph-Text Back Translator} (\textit{GT-BT}) \citep{schmitt2019unsupervised}: This approach first serializes the graph and then applies a back translation model. Since the code for GT-BT is not yet available, we adopt their reported results as a reference.
\end{itemize}

\paragraph{Supervised Baselines} 
We also compare with \textit{supervised} systems using the original supervised training data. Since there is no existing work that jointly learns graph-to-text and text-to-graph in a supervised way, we can only use models that address one of the two tasks. For graph-to-text generation, we compare with
\begin{itemize}[nolistsep]
    \item \textit{Melbourne}: The best supervised system submitted to the WebNLG challenge 2017 \citep{DBLP:conf/inlg/GardentSNP17}, which uses an encoder-decoder architecture with attention.
    \item \textit{StrongNeural} \citep{DBLP:conf/naacl/MoryossefGD19}: An enhanced version of the common encoder-decoder model.
    \item \textit{BestPlan} \citep{DBLP:conf/naacl/MoryossefGD19}: A special entity ordering algorithm applied before neural text generation.
    \item \textit{GraphWriter} \cite{DBLP:conf/naacl/Koncel-Kedziorski19}: A graph attention network with LSTMs.
    \item \textit{Seg\&Align} \citep{shen2020neural}: An approach that first segments the text into small units and then learns the alignment between data and target text segments.
    \item \textit{Planner} \cite{zhao-etal-2020-bridging}: A planner with relational graph convolutional networks and LSTMs.
    \item \textit{T5-Large} \cite{kale2020text}: An application of the pretrained T5-Large model for serialized graph-to-text generation.
    \item \textit{T5-Large} \cite{ribeiro2020investigating}: Another instance of the pretrained T5-Large model for serialized graph-to-text generation.
    \item \textit{Supervised G2T} (our implementation) : Our implementation of T5-base, which is the same as the G2T component in our \methodname{}.
\end{itemize}
For text-to-graph generation, we compare with state-of-the-art models including
\begin{itemize}[nolistsep]
    \item \textit{OnePass} \citep{DBLP:conf/acl/WangTYCWXGP19}: A BERT-based relation extraction model.
    \item \textit{T2G} (our implementation): The BiLSTM model that we adopt as the text-to-graph component in the cycle training of \methodname{}.
\end{itemize}

\subsection{Comparison Systems for GenWiki}
Since GenWiki is a non-parallel dataset, we can only train unsupervised baselines or transfer models developed using other data.  We compare against:
\begin{itemize}[nolistsep]
    \item \textit{RuleBased} \cite{schmitt2019unsupervised}: Same as was applied for WebNLG~2017 evaluation (see above).
    \item \textit{DirectTransfer}: A model trained on the supervised WebNLG~2017 dataset, and then deployed on the GenWiki test set. \item \textit{NoisySupervised}: A baseline  formed by first constructing distantly aligned pairs on the whole training data by entity overlap, and then learning from these distant supervised pairs with the G2T model.
\end{itemize}
Note that for fair comparison, we use GraphWriter \cite{DBLP:conf/naacl/Koncel-Kedziorski19} as the architecture for all G2T models including our \methodname{} on GenWiki.
\subsection{Implementation Details}

\paragraph{Processing WebNLG Data}
To test unsupervised models and baselines, we construct a non-parallel version of the training and validation sets by separating all the text in the dataset to form a text corpus, and all the graphs to build a graph dataset. We ensure that the order within the text and graph datasets are shuffled so that the data is fully non-parallel.

\paragraph{Model Details}
Our training framework can adapt to different T2G and G2T modules. For this study, we instantiate our \methodname{} framework with the pretrained T5 \cite{raffel2020exploring} as the G2T model, and a BiLSTM for the T2G architecture, with two layers comprised of 512 hidden units each. We train the model for  30 epochs.


\subsection{Main Results}

\paragraph{WebNLG~2017 Results}

As shown in Table~\ref{tab:res}, 
for G2T generation, even with no pairing information between text and graphs, our unsupervised \methodname{} model can achieve a 55.5 BLEU score, which is on par with the 56.4 BLEU score obtained using the same T2G model with T5-Base that we trained on supervised data. \methodname{} also outperforms the unsupervised baseline RuleBased and GT-BT \cite{schmitt2019unsupervised} by a clear margin.  Additionally, for the T2G task, \methodname{} also achieves scores on par with the same T2G component but trained on the supervised data. 

\paragraph{WebNLG~2020 Results}

We also submitted our system to the WebNLG~2020 Challenge. 
In Table~\ref{tab:webnlg20_g2t}, we list competing models from the competition leaderboard, as well as two baseline results published by the organizers. Since the challenge provides supervised training data, most (and possibly all) other competing systems are supervised models, such as the top-1 system, ID18 \cite{p2}, which finetunes T5 on the supervised graph-text pairs. 
As can be seen, our G2T performance is competitive with many methods even without supervision. 

\paragraph{GenWiki Results}

We show the performance of \methodname{} on the G2T task using GenWiki data in Table~\ref{tab:genwiki}.
Among the four models under consideration, 
the RuleBased and DirectTransfer methods do not perform well on the purely unsupervised dataset. The performance of DirectTransfer indicates that even though WebNLG and GenWiki are similar wiki-based datasets, it is difficult to make a model trained on one corpus to behave similarly on another slightly different one. The NoisySupervised model performs relatively well, scoring over 30 BLEU points. And our model \methodname{} is the strongest, outperforming NoisySupervised by around 10 BLEU points.

\section{Related Work}\label{sec:related_work}
We will first give an overview of the fields of T2G and T2G separately, and then introduce the cycle learning.

\paragraph{Data-to-Text Generation}

As a classic problem in text generation \citep{DBLP:conf/acl/Kukich83,DBLP:books/daglib/0073477}, data-to-text generation aims to automatically produce text from structured data \citep{DBLP:journals/nle/ReiterD97,DBLP:conf/acl/LiangJK09}.
Due to the expensive collection, all the data-to-text datasets are very small, such as the 5K air travel dataset \citep{DBLP:conf/anlp/Ratnaparkhi00}, 22K WeatherGov \citep{DBLP:conf/acl/LiangJK09}, 7K Robocup \citep{DBLP:conf/icml/ChenM08}, and 5K RotoWire on basketball games \citep{DBLP:conf/emnlp/WisemanSR17}. As for the methodology, traditional approaches adopt a pipeline system \citep{DBLP:conf/acl/Kukich83,DBLP:books/daglib/0073477} of content planning, sentence planning, and surface realization. Recent advances in neural networks give birth to end-to-end systems \citep{DBLP:conf/emnlp/LebretGA16,DBLP:conf/emnlp/WisemanSR17,DBLP:conf/naacl/Koncel-Kedziorski19,kale2020text} that does not use explicit planning but directly an encoder-decoder architecture \citep{DBLP:journals/corr/BahdanauCB14,vaswani2017nips}.

\paragraph{Relation Extraction}
Relation Extraction (RE) is the core problem in text-to-graph conversion, as its former step, entity recognition, have off-the-shelf tools with good performance \citep{lample2016neural,qian2019graphie,strakova2019neural,clark2018semi,jin2020relation}. RE aims to classify the relation of entities given a shared textual context. Conventional approaches hand-crafted lexical
and syntactic features \citep{DBLP:conf/semeval/HendrickxKKNSPP10,DBLP:conf/semeval/RinkH10}. With the recent advancement of deep neural networks, many models based on CNN \citep{DBLP:conf/coling/ZengLLZZ14,DBLP:conf/acl/SantosXZ15,DBLP:conf/naacl/NguyenG15}, RNN \citep{DBLP:conf/emnlp/SocherHMN12,DBLP:journals/corr/ZhangW15a,DBLP:conf/acl/MiwaB16,DBLP:conf/acl/ZhouSTQLHX16}, and BERT \citep{DBLP:conf/acl/WangTYCWXGP19} achieve high performance in many datasets. However, constrained by the small datasets of only several hundred or several thousand data points~\citep{walker2006ace,DBLP:conf/semeval/HendrickxKKNSPP10,DBLP:conf/semeval/GaborBSQZC18}, recent research shifts to distant supervision than model innovation \citep{DBLP:conf/acl/MintzBSJ09,DBLP:conf/emnlp/ZengLC015,DBLP:conf/acl/LinSLLS16}.


\paragraph{Cycle Training}
The concept of leveraging the transitivity of two functions inverse to each other has been widely observed on a variety of tasks.
In computer vision, the forward-backward consistency has been used since last decade \citep{DBLP:conf/icpr/KalalMM10,DBLP:conf/eccv/SundaramBK10}, and training on cycle consistency has recently been extensively applied on image style transfer \citep{DBLP:conf/iccv/ZhuPIE17,DBLP:conf/cvpr/GodardAB17}. In language, back translation \citep{DBLP:conf/acl/SennrichHB16,DBLP:conf/emnlp/EdunovOAG18,jin2020simple} and dual learning \citep{DBLP:conf/acl/ChengXHHWSL16,DBLP:conf/nips/HeXQWYLM16} have also been an active area of research centered on UMT. Similar techniques can also be seen on tasks such as language style transfer \citep{DBLP:conf/nips/ShenLBJ17,DBLP:conf/emnlp/JinJMMS19}. Finally, cycle training has recently been applied in the context of text-graph conversion \cite{schmitt2019unsupervised}; we compare against this approach in Table~\ref{tab:res} (see GT-BT results).

\section{Conclusion}\label{sec:future}

We have developed a cycle learning framework for both text-to-graph and graph-to-text generation in an unsupervised way. Experimental results validate that the proposed model achieves comparable results to many supervised models, given the same amount of unsupervised data. Moreover, when comparing with other unsupervised models, our \methodname{} model displays a clear advantage.

\section*{Acknowledgements}
We thank colleagues at the Amazon Shanghai AI lab, including Xiangkun Hu, Hang Yan, and many others for insightful discussions that constructively helped this work. 

\bibliography{anthology,emnlp2020}
\bibliographystyle{acl_natbib}

\appendix

\end{document}